\documentclass[10pt,twocolumn,letterpaper]{article}

\usepackage[pagenumbers]{cvpr} 
\usepackage{makecell}
\definecolor{cvprblue}{rgb}{0.21,0.49,0.74}
\usepackage[pagebackref,breaklinks,colorlinks,allcolors=cvprblue]{hyperref}
\usepackage{amsmath}

\def\paperID{} 
\def\confName{CVPR}
\def\confYear{2026}

\title{MoCha: End-to-End Video Character Replacement without Structural Guidance}

\author{
  Zhengbo Xu, 
  Jie Ma, 
  Ziheng Wang, 
  Zhan Peng, 
  Jun Liang,
  Jing Li
  \vspace{6pt}\\
  \vspace{6pt}
  Project Page: \url{orange-3dv-team.github.io/MoCha}
}
\begin{document}

\twocolumn[{%
\renewcommand\twocolumn[1][]{#1}%
\maketitle      
\vspace{-8mm}
\begin{center}
    \centering
    \includegraphics[width=\linewidth]{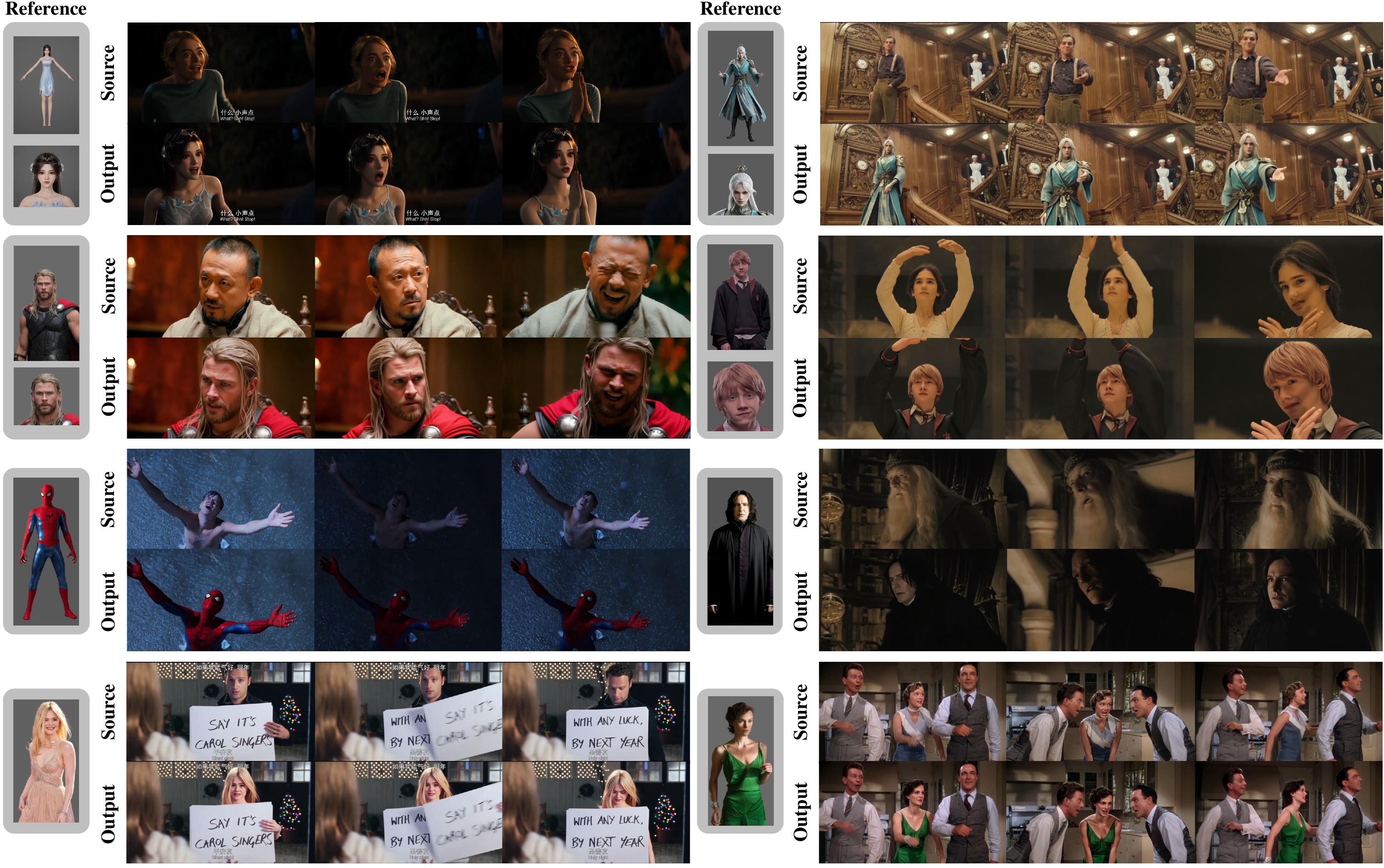}
    \vspace{-6mm}
    \captionof{figure}{
    \textbf{Examples generated by MoCha.} MoCha enables high-fidelity character replacement in source videos based on provided reference images for diverse subjects, including virtual (first row) and real-human (second row) characters. Furthermore, our approach robustly preserves original lighting conditions (third row) and effectively handles multi-character occlusion and interaction (fourth row).
    } \vspace{1mm}
    \label{fig:teaser}
\end{center}
}]

\begin{abstract}
Controllable video character replacement with a user-provided identity remains a challenging problem due to the lack of paired video data.
Prior works have predominantly relied on a reconstruction-based paradigm that requires per-frame segmentation masks and explicit structural guidance (e.g., skeleton, depth). 
This reliance, however, severely limits their generalizability in complex scenarios involving occlusions, character-object interactions, unusual poses, or challenging illumination, often leading to visual artifacts and temporal inconsistencies.
In this paper, we propose MoCha, a pioneering framework that bypasses these limitations by requiring only a single arbitrary frame mask. 
To effectively adapt the multi-modal input condition and enhance facial identity, we introduce a condition-aware RoPE and employ an RL-based post-training stage.
Furthermore, to overcome the scarcity of qualified paired-training data, we propose a comprehensive data construction pipeline. Specifically, we design three specialized datasets: a high-fidelity rendered dataset built with Unreal Engine 5 (UE5), an expression-driven dataset synthesized by current portrait animation techniques, and an augmented dataset derived from existing video-mask pairs.
Extensive experiments demonstrate that our method substantially outperforms existing state-of-the-art approaches.
We will release the code to facilitate further research.
Please refer to our project page for more details: \url{orange-3dv-team.github.io/MoCha}

\end{abstract}    
\section{Introduction}
Recent breakthroughs in generative technologies, particularly diffusion models~\cite{flux2024, batifol2025flux, wan2025, kong2024hunyuanvideo}, have propelled content creation into a new era. Consequently, user demand for fine-grained and highly personalized editing of images and videos surges rapidly. Video character replacement~\cite{WanAnimate}, defined as the task of seamlessly substituting the character in a video while precisely preserving the original background, scene dynamics, and character motion, demonstrates substantial practical value and growing commercial significance. Its applicability spans diverse domains, including costly post-production in film and television, personalized advertising, virtual try-on, and the creation of dynamic digital avatars.

\begin{figure}
    \centering
    \includegraphics[width=\linewidth]{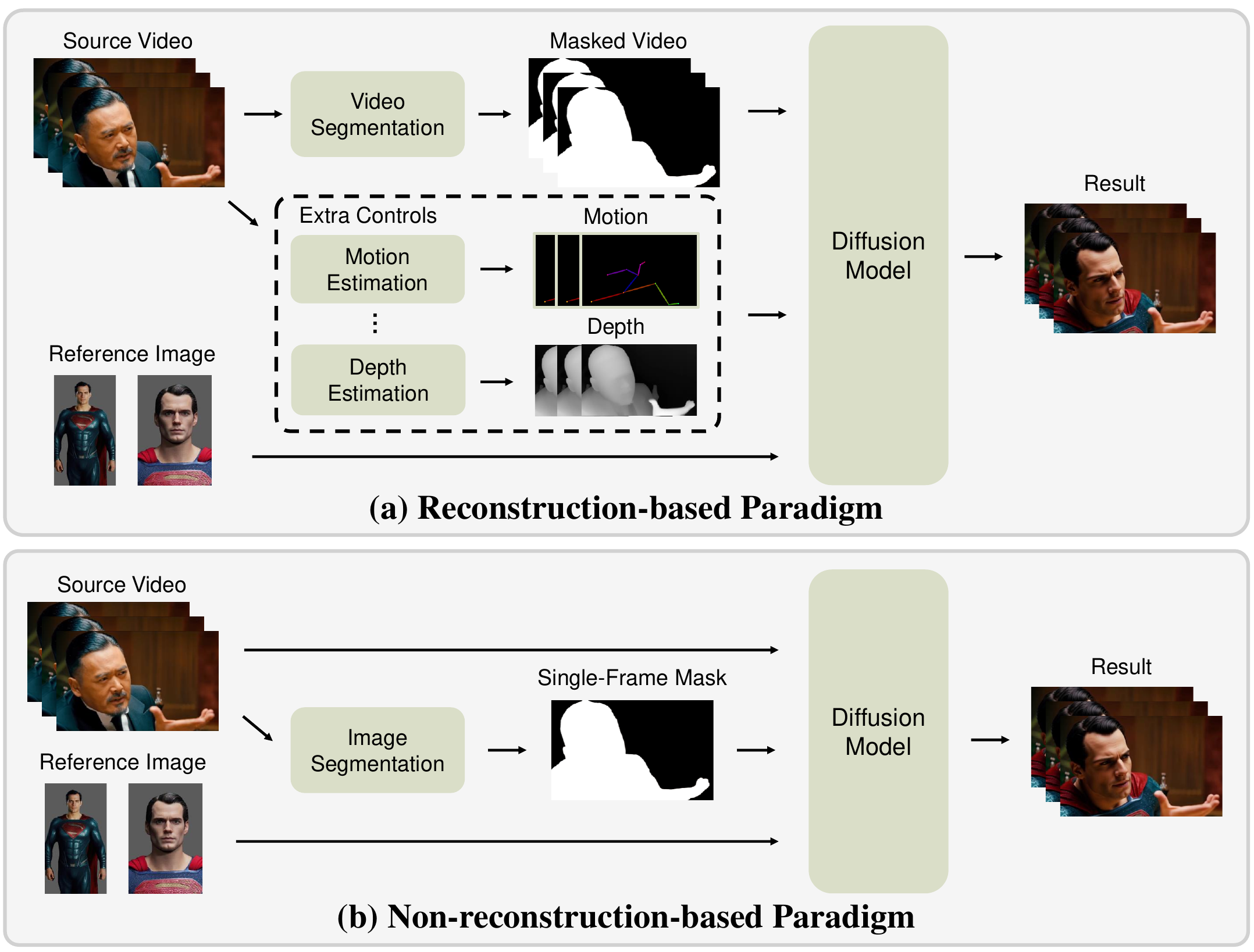}
    \vspace{-6mm}
    \captionof{figure}{
    (a) Reconstruction-based paradigm used in baseline methods. (b) Non-reconstruction-based paradigm used in MoCha.} \vspace{-1.5em}
    \label{fig:reconstruct}
\end{figure}

Despite its significance, video character replacement still faces significant challenges. Current research~\cite{jiang2025vace, WanAnimate, hu2025hunyuancustom} has predominantly adopted a reconstruction-based paradigm, as shown in Fig.~\ref{fig:reconstruct}. These methods first require a dense, per-frame segmentation mask to annotate the character spatial location and ruin its ID information. 
Then they utilize the structural guidance, such as pose skeletons~\cite{xu2022vitpose} or depth maps~\cite{yang2024depth} extracted from the source video, along with a reference image of the new character, to re-render the video.
While this paradigm produce impressive results in simple scenarios, it struggles in complex ones—such as videos with occlusion, unusual poses (e.g., acrobatics), and multi-character interactions involving physical contact.
In these cases, the multi-frame masks and structural information are prone to inaccuracies.
Consequently, inaccuracies in these explicit guidance signals are further propagated and amplified during the generation process, resulting in severe visual artifacts, motion discontinuities, and temporal inconsistency in the rendered videos.
Furthermore, the reliance on dense explicit guidance not only limits the flexibility and robustness of video character replacement algorithms but also incurs substantial computational overhead.

Recent advances have shown that video diffusion models inherently perform temporal perception and implicit reasoning abilities~\cite{veo3reason}.
Harnessing these capabilities, specifically video tracking, we challenge the reliance on a per-frame mask for this task.
In this paper, we propose MoCha, the first end-to-end framework for video character replacement that requires only a single frame mask from an arbitrary video frame without any structural guidance. 
MoCha operates by decoupling the source character's motion and facial expressions from the background scene, and subsequently transferring these dynamics onto a new reference identity through efficient in-context learning by integrating the video content, frame mask, and reference character identity.
To coherently fuse these multi-modal inputs, we propose a condition-aware RoPE, an extension of the original 3D RoPE~\cite{rope, wan2025} mechanism.
Furthermore, to better preserve the character's identity, we introduce a post-training stage guided by a differentiable facial reward function~\cite{deng2019arcface}.
Benefiting from these designs, MoCha demonstrates strong potential in achieving accurate and temporally consistent video character replacement.

Training MoCha requires a high-quality paired dataset consisting of source videos and corresponding target videos, in which the character is replaced while preserving the character motion and scene dynamics. 
To this end, we introduce a comprehensive data construction pipeline that aggregates data from three distinct sources, as shown in Fig.~\ref{fig:data}. 
First, we use Unreal Engine 5 (UE5) to render paired character video, ensuring that different characters perform identical actions within the same scene. 
Second, we generate animated portrait videos by first replacing the person in a source image using the Flux inpainting model ~\cite{flux2024}, and then animating both the original and inpainted images using current portrait animation methods~\cite{liveportrait} with a shared driving facial expression sequence. 
Third, to further enrich data diversity, we incorporate publicly available video-mask datasets~\cite{hu2024vivid}.
By training on this composite dataset, MoCha can perform video character replacement in an end-to-end manner, without requiring auxiliary inputs like per-frame masks or structural guidance.
Our contributions can be summarized as follows:
\begin{figure*}
    \centering
    \includegraphics[width=\linewidth]{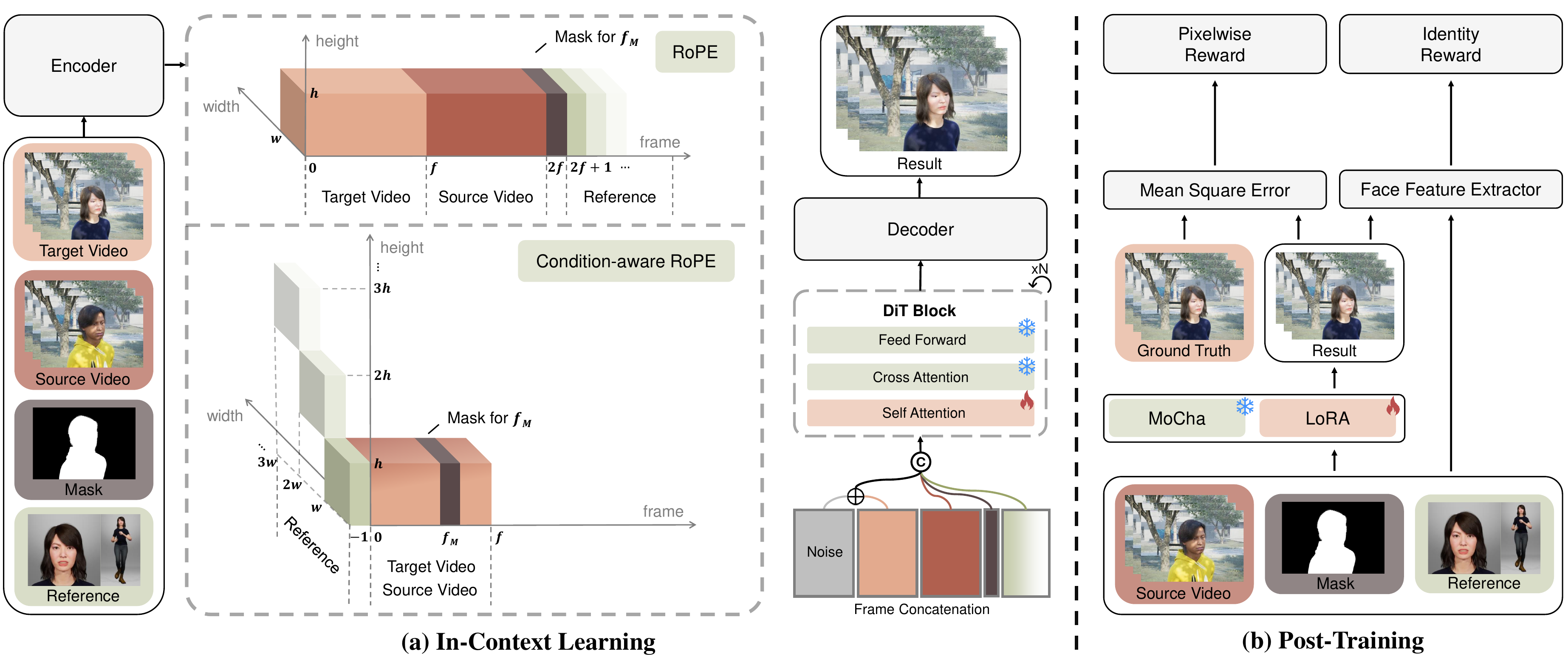}
    \vspace{-6mm}
    \caption{
    \textbf{Overview of MoCha.} Training MoCha consists of two stages: (a) In-Context Learning. We apply the condition-aware RoPE to the concatenated tokens and train the DiT backbone. (b) Post-Training. We employ an RL-based strategy to further enhance the facial consistency. }
    \vspace{-1.5em}
    \label{fig:method}
\end{figure*}

\begin{itemize}
    \item We introduce a high-quality and large-scale paired dataset for video character replacement.

    \item We propose MoCha, the first end-to-end framework for video character replacement that requires only a single frame mask from an arbitrary source video frame without any structural guidance. MoCha also demonstrates the tracking ability of the video diffusion model.

    \item Extensive experiments demonstrate that MoCha significantly outperforms current state-of-the-art methods in terms of identity preservation, temporal consistency, and facial expression fidelity. 
\end{itemize}

\section{Related Works}
\textbf{In-Context Learning in Video Generation.}
Recent breakthroughs in video generation have ushered in a new era for content creation. 
State-of-the-art video diffusion models~\cite{wan2025, kong2024hunyuanvideo} are now capable of producing high-fidelity, long-duration videos. 
Building on these foundation models, various works explore in-context learning for different downstream tasks. 
This approach typically involves guiding the generation process by concatenating conditioning inputs along the temporal dimension. 
For instance, Recammaster~\cite{Recammaster} frames camera-controlled video re-rendering as an in-context learning task, FullDiT~\cite{ju2025fulldit} integrates multiple conditions for versatile video generation, and VFXMaster~\cite{li2025vfxmaster} applies this method to transfer visual effects from a video to an image.
Motivated by the demonstrated effectiveness of this paradigm, our work also adopts an in-context learning framework.

\medskip
\noindent\textbf{Video Character Replacement.}
Character replacement in video is a key challenge in video editing, with recent methods showing significant progress. A dominant paradigm among current methods is reconstruction-based synthesis, where a specified character is generated within a masked region of a target video. 
HunyuanCustom~\cite{hu2025hunyuancustom} simply masks the target area and concatenates the reference image.
VACE~\cite{jiang2025vace} incorporates structural guidance, such as depth or skeleton maps, into the DiT diffusion model to maintain character movement.
Building upon this, Wan-Animate~\cite{WanAnimate} introduces a face encoder to extract fine-grained detail to maintain facial expressions.
However, a key limitation of these approaches is their reliance on dense, per-frame segmentation masks and structural priors.
In contrast, our method, MoCha, achieves end-to-end character replacement via an implicit learning mechanism. It requires only a single frame mask of the video, thus obviating the need for such explicit guidance and streamlining the synthesis process.
\vspace{-1em}
\section{Method}

MoCha is designed for the video character replacement task, which requires a source video $V_s \in \mathbb{R}^{f \times c \times h \times w}$, a designation frame mask $M_s \in \mathbb{R}^{1 \times 1 \times h \times w}$ to annotate the character and a set of refernece character image $\{I_i \in \mathbb{R}^{1 \times 1 \times h \times w}\}$, as shown in Fig.~\ref{fig:method}.
In this section, we first review the Flow matching framework that is commonly used in current video diffusion models in Sec.\ref{sec:Preliminary}. Then we introduce our in-context learning framework in Sec.\ref{sec:in-context} and reward post-training in Sec.\ref{sec:post_training}.

\subsection{Preliminary}
\label{sec:Preliminary}
MoCha builds on the pretrained text-to-video latent diffusion model using the Rectified Flow framework~\cite{flow-matching}. The input video $V_0$ will first be compressed to a latent $z_0$ through a video variational encoder~\cite{vae} (VAE). Based on the clean latent $z_0$, a noisy latent $z_t$ is constructed by a randomly sampled timestep $t$ and a noise $\epsilon$ sampled from standard Gaussian Distribution $\mathcal{N}(0,1)$:
\begin{equation}
z_t = (1-t)z_0 + t\epsilon
\end{equation}

During training, $z_t$ is fed to a Transformer-based diffusion model~\cite{Peebles2022DiT} (DiT). The model predicts the velocity $v_{\Theta}(z_t, t)$ at point $z_t$, the gradient value satisfying $\mathrm{d}z_t = v_{\Theta}(z_t, t) \mathrm{d}t$. We use the Conditional Flow Matching~\cite{CFM} loss to optimize the model:
\begin{equation}
\mathcal{L}_{FM} = \mathbb{E}_{z_0,t,\epsilon} || v_{\Theta}(z_t, t) - u_t(z_0 | \epsilon)||_2
\end{equation}
where $u_t(z_0 | \epsilon)$ is the actual slope of the secant line through $z_0$ and $\epsilon$. For inference, we start from a pure noise $z_1$ and iteratively generate the output latent $z_0$ through a timestep list $\{t_i\}$:
\begin{equation}
z_{t_{i+1}} = z_{t_i} + v_{\Theta}(z_{t_i}, t_i) \cdot (t_{i+1} - t_i)
\end{equation}
where $\{t_i\}$ is a decreasing list with $t_0 = 1$ and $t_n = 0$. $z_0$ is finally decoded to the output video.

\subsection{In-Context Learning}
\label{sec:in-context}
To guide the generation of a target video $V_t$ based on a source video $V_s$, a designation mask $M$ for a single frame, and a set of reference images $\{I_i\}$, we employ an in-context learning manner due to its effectiveness demonstrated in various tasks~\cite{ju2025fulldit,he2025fulldit2,Recammaster,li2025vfxmaster}. 
Concretely, we first encode these conditions with a 3D VAE $\mathcal{E}$ to achieve spatiotemporal compression.
This yields the latents $z_t = \mathcal{E}(V_t)$, $z_s = \mathcal{E}(S)$, $z_m = \mathcal{E}(M)$, $z_{I_i} = \mathcal{E}(I_i)$.
These latents are then patchified into visual tokens, namely $x_t \in \mathbb{R}^{b\times f\times c \times h \times w}$, $x_s \in \mathbb{R}^{b\times f\times c \times h \times w}$, $x_m\in \mathbb{R}^{b\times 1\times c \times h \times w}$, and $x_{I_i}\in \mathbb{R}^{b\times 1\times c \times h \times w}$.
Here $b$, $f$, $c$, $h$, $w$ represent the batch size, frame count, channel number, height, and width, respectively.
Next, we concatenate all conditional tokens with the target tokens along the frame dimension to form a unified latent sequence:
\begin{equation}
    x = [x_t, x_s, x_m, x_{I_1}, x_{I_2}, ...]
\end{equation}
where $x \in \mathbb{R}^{b \times (2f+1+j) \times c \times h \times w}$ and $j$ is the number of reference images.
The latent sequence $x$ is subsequently fed into the DiT backbone and processed using full self-attention.

However, a naive application of 3D Rotary Positional Embeddings (RoPE) by assigning a different temporal index to each condition would result in inflexible generation, such as being constrained to a fixed output length. 
To address this, we propose a condition-aware RoPE mechanism to coherently fuse these diverse inputs, as shown in Fig.~\ref{fig:method}. 
Concretely, we assign the same frame index, ranging from $0$ to $f-1$, to both the source video tokens $x_s$ and target video tokens $x_t$, as they share a frame-to-frame correspondence. 
For the reference image tokens ${I_i}$, a fixed frame index of $-1$ is assigned, and different reference image tokens are distinguished using an offset in the height and width dimensions.
A key innovation of MoCha is that it supports flexible arbitrary frame mask selection, which is implemented by assigning a variable frame index $f_M$ for the mask token $x_m$ based on the designation frame number $F$, i.e.
\begin{equation}
    f_M = (F-1) // 4 + 1
\end{equation}
This in-context learning manner, equipped with our designed condition-aware RoPE, enables MoCha to effectively handle
variable generation length, flexible multi-reference images, and arbitrary frame mask selection.

\subsection{Identity-Enhancing Post-Training}
\label{sec:post_training}
Recent works \cite{liang2025aesthetic, shen2025directly} have explored diffusion post-training with reinforcement learning to align human preferences.
Drawing inspiration from this, we employ a similar RL-based post-training strategy to further enhance the facial consistency between the generated video and the reference images, as shown in Fig. \ref{fig:method} (b).
The core idea is to guide the generation process through an explicit optimization objective that maximizes facial consistency between the generated video frames and ground-truth video frames.
Specifically, we compute a facial reward score $R_{face}$ based on the cosine similarity between the Arcface~\cite{deng2019arcface} embeddings of the generated video and the reference images.
To avoid reward hacking that the model simply pastes the reference image into the generated video, we also use a pixel-wise Mean Squared Error (MSE) loss between the generated video and the GT video to provide dense supervision. The overall loss function is:
\begin{equation}
    \mathcal{L}_{RL} = (1-R_{face}) + ||V_t - \hat{V_t}||_2
\end{equation}
where $\hat{V_t}$ is the video generated by the model.
Since fine-grained details are primarily synthesized in the later timestamps of the sampling process~\cite{prabhudesai2023aligning}, we only backpropagate the loss through the final $K$ sampling steps to reduce memory and accelerate optimization.

\begin{figure*}
    \centering
    \includegraphics[width=\linewidth]{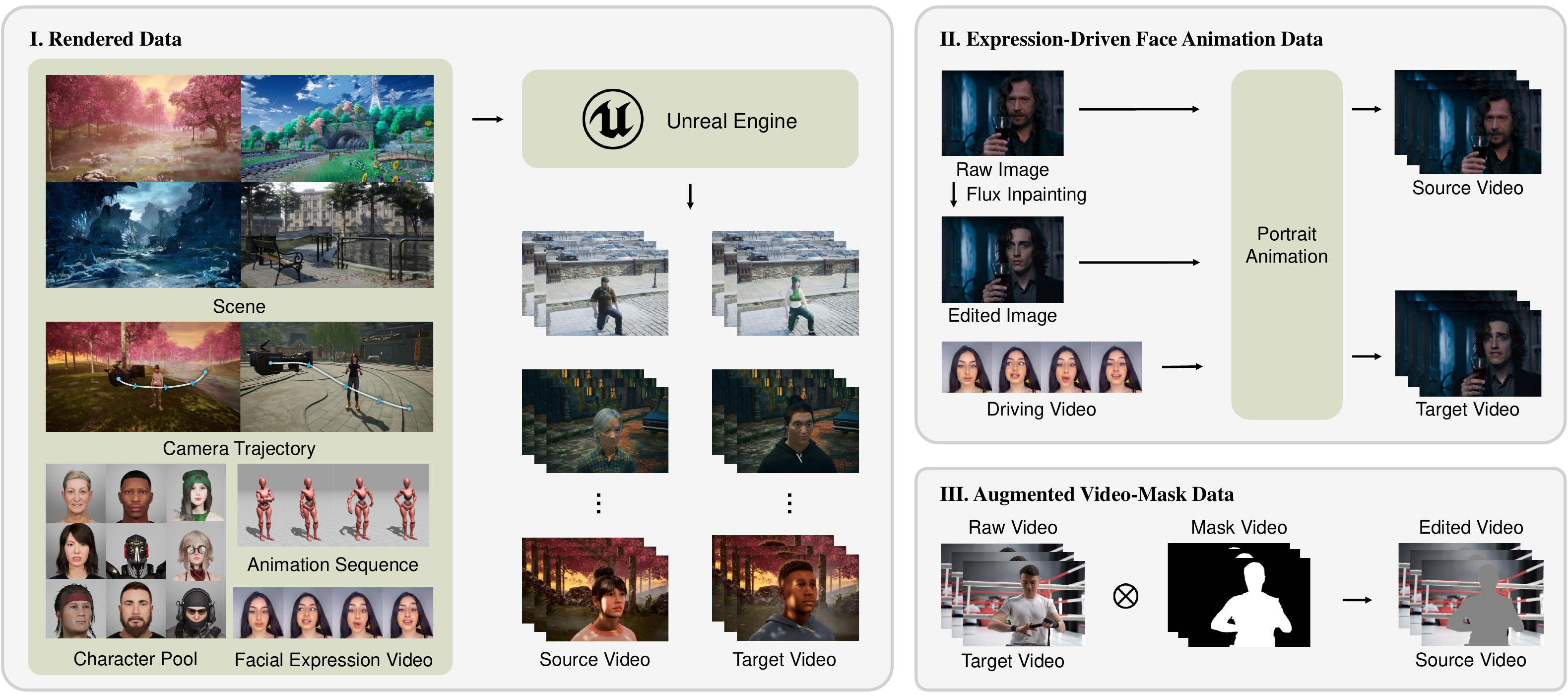}
    \vspace{-6mm}
    \caption{
    \textbf{Overview of the data construction pipeline.} We propose three methods to construct the training dataset: (I) Rendered data built with Unreal Engine 5. (II) Expression-driven face animation data generated through portrait animation methods. (III) Augmented data synthesized from traditional video-mask pairs.}
    \vspace{-1.5em}
    \label{fig:data}
\end{figure*}
\section{Dataset}
\label{sec:dataset}
Training MoCha requires strictly aligned paired video, where different characters share the same motion, facial expressions and background dynamics. However, it is challenging to obtain such pairs in real-world settings. To address this critical limitation, we introduce a comprehensive data construction pipeline that aggregates data from three distinct sources. This process is illustrated in Fig.~\ref{fig:data}.

\subsection{Rendered Data} 
We develop a scalable data generation pipeline built upon Unreal Engine 5 (UE5) to synthesize a large-scale dataset of paired videos. The pipeline leverages an extensive library of virtual assets from the UE ecosystem, including 3D scenes, characters, motions, and facial expressions.
Each video is rendered through the random composition of these assets.
To enrich the dataset with diverse perspectives, we also implement a procedural system to automatically generate natural camera trajectories.

The core of our dataset lies in its paired structure. For each video, a corresponding paired video is rendered by substituting the character while meticulously preserving all other parameters.
Additionally, a video character mask is also rendered to precisely annotate the replaced character location.
Finally, for each character, a set of reference images is rendered, capturing them in various poses and under different lighting conditions.

\subsection{Expression-Driven Face Animation Data}
\label{sec: expression}
To enhance facial expression fidelity, we further construct an expression-driven paired dataset by the current portrait animation methods. We first collect a large set of film images and use the Flux inpainting model \cite{flux2024} to substitute the foreground character. Then we use LivePortrait \cite{liveportrait} to animate both images with the same facial-driven video. 
Simply using an image from the video as a reference image may cause a copy-paste problem, i.e., the model tends to directly paste the reference image to the generated video.
Therefore, instead of using a raw frame from the driving video as the reference, we augment its pose with Flux Kontext \cite{batifol2025flux}. This creates a novel reference that forces the model to decouple identity from facial animation. 

\subsection{Augmented Video-Mask Data}
Training MoCha solely on the two aforementioned synthetic datasets may lead to noticeable synthetic artifacts and a lack of realism in the generated characters. 
To address this, we enrich our training set with real-world videos from two public datasets that provide video-mask pairs: VIVID-10M~\cite{hu2024vivid} and VPData~\cite{bian2025videopainter}. 
We apply a YOLOv12 detector \cite{tian2025yolov12} to filter non-human videos. 
The references are augmented using the similar method mentioned in Sec~\ref{sec: expression}.

\section{Experiments}

\begin{figure*}
    \centering
    \includegraphics[width=\linewidth]{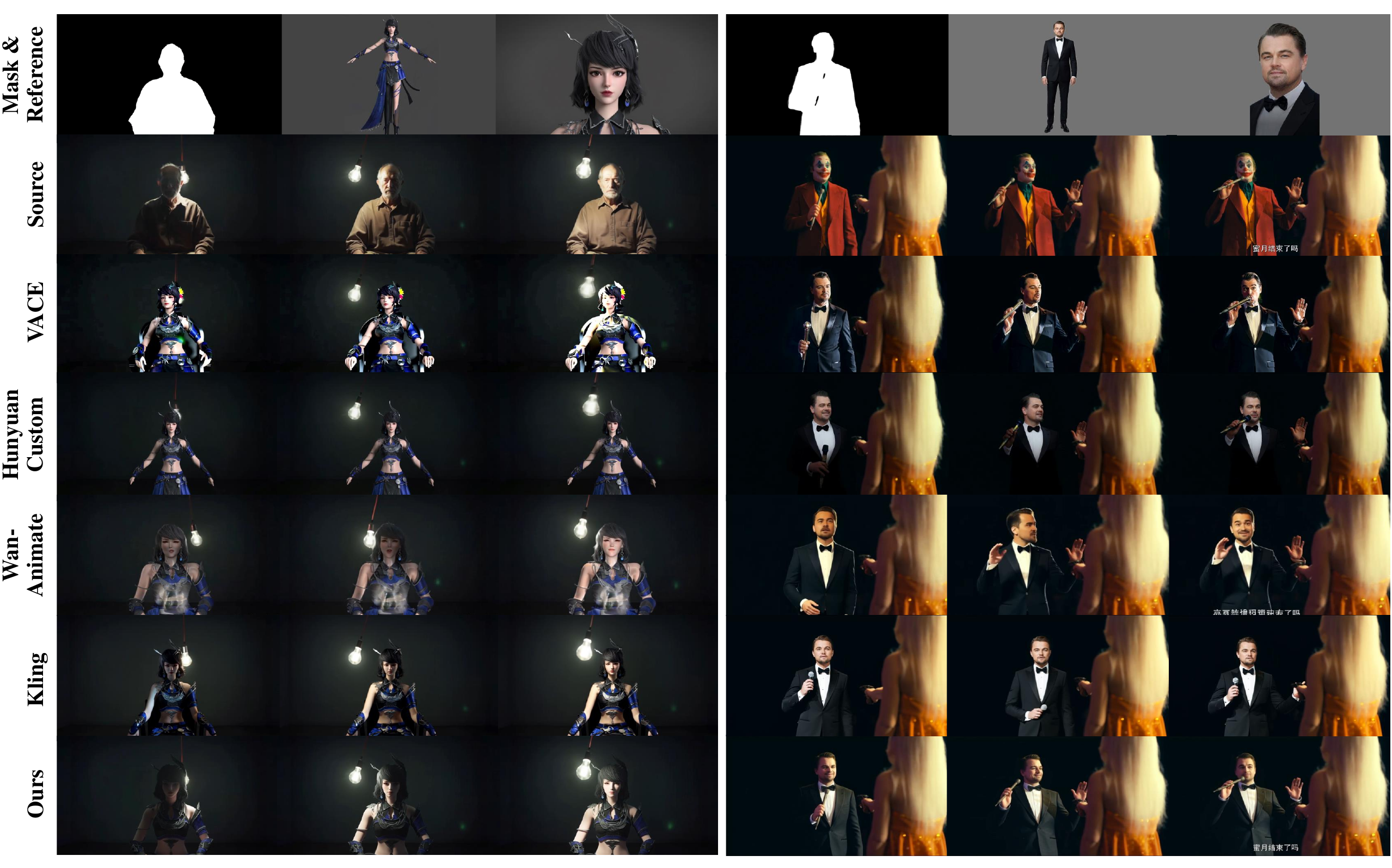}
    \caption{
    \textbf{Comparison with state-of-the-art methods.} The results show that MoCha can replace the character with more consistent animation, higher facial expressiveness, and more natural lighting effects.}
    \label{fig:comparison}
    \vspace{-1.5em}
\end{figure*}

\subsection{Experiment Settings}
\textbf{Implementation Details.}~
We use Wan-2.1-T2V-14B~\cite{wan2025} as the base model, due to its excellent performance. The training process consists of two stages. In the in-context learning stage, we fine-tune all self-attention layers to handle multi-model input conditions. Training is conducted on $8$ NVIDIA H20 GPUs for $30$K steps. We use a learning rate of $2 \times 10^{-5}$, a batch size of $8$. In the post-training stage, to avoid rewards affecting the generation ability of the base model, we employ parameter-efficient fine-tuning using Low-Rank Adaptation (LoRA)~\cite{hu2022lora} instead of full model tuning. We apply LoRA with a rank of $64$ to all linear layers within the DiT architecture. The LoRA modules are then trained for $500$ steps, using a batch size of $8$ and a learning rate of $2 \times 10^{-5}$.

Our dataset consists of three distinct sources. We randomly select 60K, 20K, 20K videos from rendered data, expression-driven animation data, and augmented video-mask data, respectively, in a total of 100K samples. All videos, masks, and reference images are resized to $480 \times 832$ resolution for training.

\medskip
\noindent\textbf{Training Strategy.}~
MoCha supports multiple reference images as input. During training, we always provide one base reference and, with 50\% probability, include an additional face-centric reference. This stochastic reference-dropout scheme improves robustness when reference cues are limited and enhances the model’s ability to exploit extra references to learn fine-grained facial details. In addition, we initially train the model using short video clips (e.g., 21 frames). Once the generation results stabilize, we increase the length of the input video sequences (e.g., 81 frames) in subsequent training stage. 

\medskip
\noindent\textbf{Benchmark.}~
To facilitate a comprehensive quantitative evaluation, we introduce a new benchmark composed of two distinct subsets: a synthetic benchmark and a real-world benchmark.
The synthetic benchmark is constructed by a render engine, providing perfectly paired data. To ensure a fair comparison, none of the scenes, characters, motions, or facial expressions in the benchmark appear in our training data. For evaluation on this benchmark, we employ widely-used quantitative metrics, including SSIM~\cite{wang2004ssim}, LPIPS~\cite{zhang2018lpips}, and PSNR.

The real-world benchmark consists of a diverse set of videos collected to encompass challenging scenarios, such as multi-person interactions, rapid movements, and complex lighting conditions.
For each video, we use SAM2~\cite{sam2} to generate a randomly selected frame mask.
We also curate a diverse collection of reference images, spanning various artistic styles (e.g., cartoon, photorealistic) and subject attributes (e.g., gender, body shape).
To assess performance on this benchmark, we adopt the comprehensive evaluation suite from VBench~\cite{huang2024vbench} for a thorough comparison.

\subsection{Comparison on Character Replacement}
\textbf{Baselines.}~We compare MoCha with the current state-of-the-art multimodal video editing methods including VACE~\cite{jiang2025vace}, Kling~\cite{kling}, Wan-Animate~\cite{WanAnimate} and HunyuanCustom~\cite{hu2025hunyuancustom}. Among them, Kling~\cite{kling} is a commercial product supporting online character replacement. Other methods are open-sourced and we use their official implementations for evaluation.

\medskip
\noindent\textbf{Qualitative Results.}~
We present several character replacement results generated by MoCha in Fig.~\ref{fig:teaser}. Our model demonstrates robust performance across diverse subject styles, handling both cartoon and real-human characters. Specifically, MoCha is capable of reproducing vivid facial expressions and transferring complex animation onto the reference characters.

We further compare MoCha with state-of-the-art methods in Fig.~\ref{fig:comparison}. While all these methods can generate relatively high-quality video based on a reference character, they exhibit distinct limitations. Specifically, Kling~\cite{kling} and HunyuanCustom~\cite{hu2025hunyuancustom} successfully preserve the characteristics of the reference character, but fail to maintain the character's original motion and cannot integrate the character naturally into the source video.
Wan-Animate~\cite{WanAnimate} and VACE~\cite{jiang2025vace} preserve the clothing of target character well. However, they struggle to learn precise lighting and shading information because their reconstruction-based paradigm loses a significant amount of original video information.
In comparison, MoCha can replace the specified character with high identity and animation consistency, all while preserving the original environment details and lighting effects.

\medskip
\noindent\textbf{Quantitative Results.}~
We compare MoCha with other methods on the aforementioned quantitative metrics. It is important to note that we omit Kling from the comparison because it lacks the capability for large-scale, automated batch testing.
As shown in Tab.~\ref{tab:quantitative}, MoCha consistently achieves state-of-the-art performance across all reference-based metrics, demonstrating superior temporal coherence and structural fidelity.
For VBench \cite{huang2024vbench} metrics, we select six different evaluation dimensions most closely related to our character replacement task. Tab.~\ref{tab:vbench} illustrates that videos generated by MoCha exhibit superior quality and consistency compared to existing methods.

\begin{table}[htbp]
  \centering
  \caption{Quantitative comparison with state-of-the-art methods on synthesized benchmark on structural similarity (SSIM), perceptual similarity (LPIPS) and reconstruction quality (PSNR).}
  \label{tab:quantitative}
  \vspace{-3mm}
  \begin{tabular}{lcccc}
    \toprule
    \textbf{Method} & \textbf{SSIM$\uparrow$} & \textbf{LPIPS$\downarrow$} & \textbf{PSNR$\uparrow$} \\
    \midrule
    VACE & 0.572 & 0.253 & 17.10 \\
    HunyuanCustom & 0.644 & 0.257 & 17.70 \\
    Wan-Animate & 0.692 & 0.213 & 19.20 \\
    \midrule
    \textbf{MoCha} & \textbf{0.746} & \textbf{0.152} & \textbf{23.09} \\
    \bottomrule
  \end{tabular}
  \vspace{-1.5em}
\end{table}

\begin{table*}[htbp]
  \centering
  \caption{Quantitative comparison with state-of-the-art methods on real-world benchmark on VBench~\cite{huang2024vbench} metrics.}
  \label{tab:vbench}
  \vspace{-3mm}
  \begin{tabular}{lcccccc}
    \toprule
    \textbf{Method} & 
    \makecell{\textbf{Subject} \\ \textbf{Consistency} $\uparrow$} & 
    \makecell{\textbf{Background} \\ \textbf{Consistency} $\uparrow$} &
    \makecell{\textbf{Aesthetic} \\ \textbf{Quality} $\uparrow$} & 
    \makecell{\textbf{Imaging} \\ \textbf{Quality} $\uparrow$} & 
    \makecell{\textbf{Temporal} \\ \textbf{Flickering} $\uparrow$} & 
    \makecell{\textbf{Motion} \\ \textbf{Smoothness} $\uparrow$}
     \\
    \midrule
    VACE & 71.19 & 77.89 & 56.76 & \textbf{60.88} & 97.04 & 97.87 \\
    HunyuanCustom & 90.03 & 93.68 & 56.77 & 58.92 & \textbf{97.98} & 98.62 \\
    Wan-Animate & 91.25 & 93.42 & 54.60 & 58.48 & 97.27 & 98.25 \\
    \midrule
    \textbf{MoCha} & \textbf{92.25} & \textbf{94.40} & \textbf{60.24} & 59.58 & \textbf{97.98} & \textbf{98.79} \\
    \bottomrule
  \end{tabular}
  \vspace{-1.5em}
\end{table*}

\subsection{Ablation Study}
\begin{figure}
    \centering
    \includegraphics[width=\linewidth]{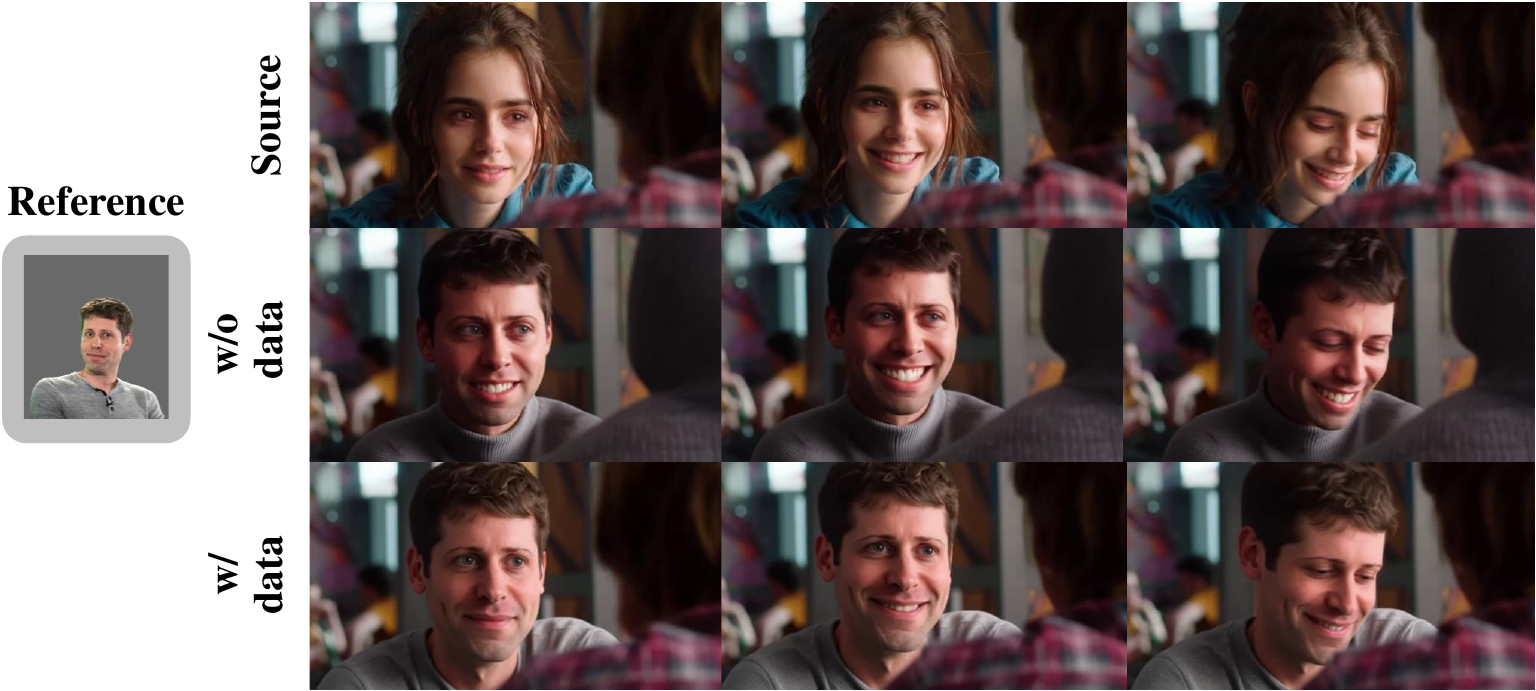}
    \captionof{figure}{
    \textbf{Ablation on Real-Human Data.} The result shows that real-human data improves the overall realism and facial fidelity of the output.}
    \label{fig:ablation_data}
    \vspace{-1em}
\end{figure}
\textbf{Ablation on Real-Human Data.}
In this paper, we propose several methods to construct paired training data for the character replacement task. Among these, the rendered dataset forms the foundation of our training data since its paired video is strictly aligned. To validate the contribution of the remaining real-human datasets (i.e., the expression-driven data and the augmented video-mask data), we compare the model's performance when trained with and without their inclusion.

As illustrated in Fig.~\ref{fig:ablation_data}, the real-human datasets significantly enhance the generated character's facial appearance, yielding expressions that more accurately align with the source video character. Furthermore, the result indicates that these data are crucial for overall quality: they effectively mitigate the synthetic artifacts (e.g., an overly smooth or glossy texture) brought by the rendered data, and substantially generate more realistic and higher-fidelity characters.

\medskip
\begin{figure}
    \centering
    \includegraphics[width=\linewidth]{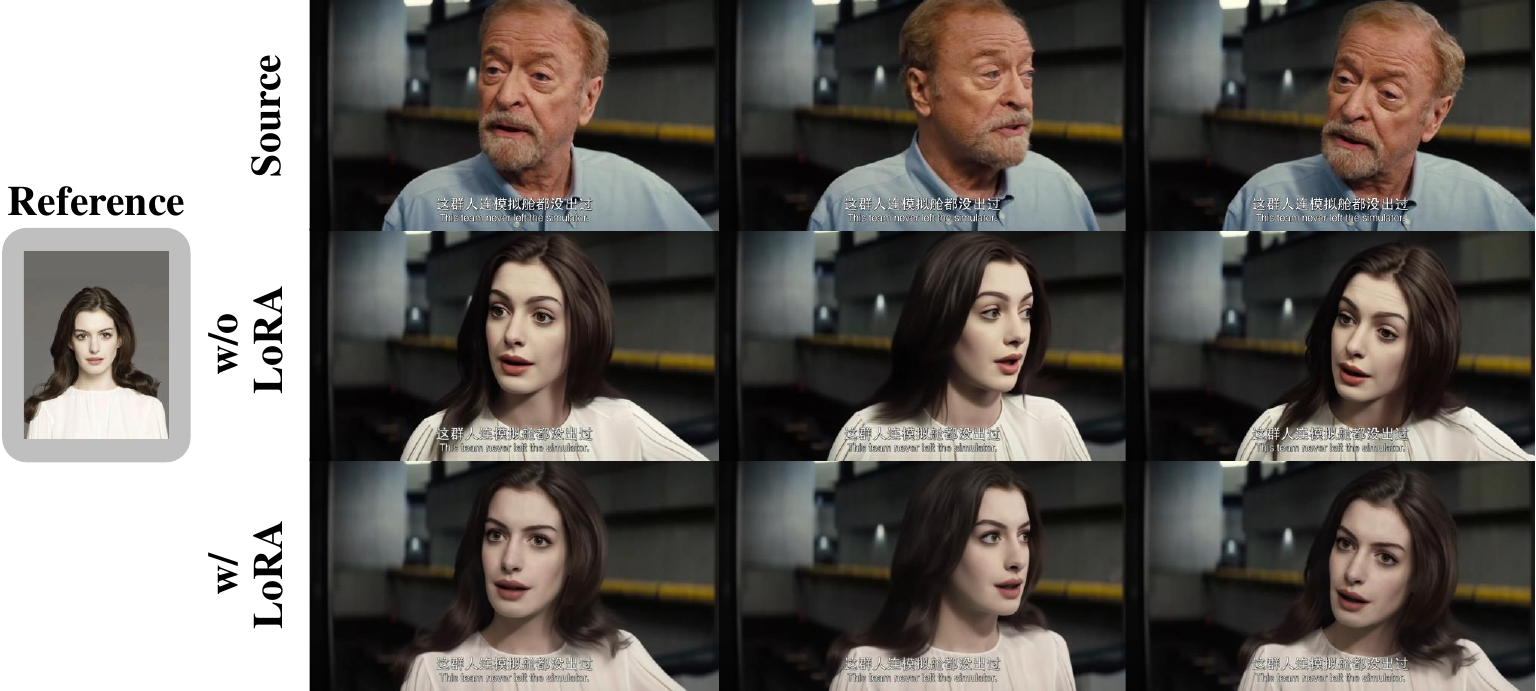}
    \captionof{figure}{
    \textbf{Ablation on Identity-Enhancing Post-Training.} The result shows that our post-training strategy is crucial to enhance the facial consistency.}
    \label{fig:ablation_lora}
    \vspace{-1.5em}
\end{figure}
\noindent\textbf{Ablation on Identity-Enhancing Post-Training.}
In Section~\ref{sec:post_training}, we introduce an RL-based identity-enhancing post-training strategy to overcome the challenge of imperfect identity consistency in the face of the generated video relative to the reference image. 
As shown in Fig.~\ref{fig:ablation_lora}, equipped with reward LoRA, our model significantly improves the character's facial identity preservation while simultaneously maintaining the original generation quality. 

\subsection{More Analysis}
\begin{figure}
    \centering
    \includegraphics[width=\linewidth]{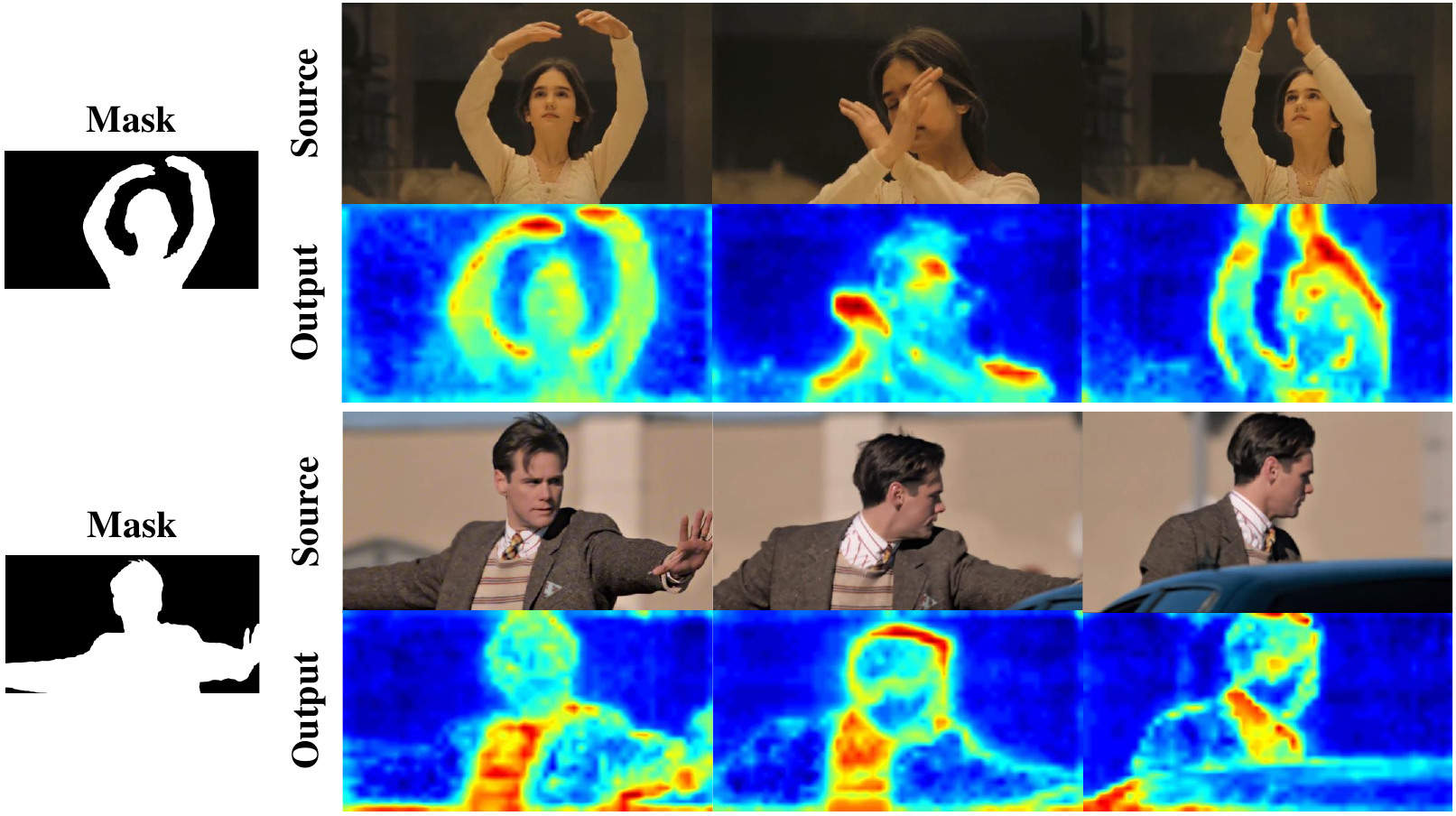}
    \vspace{-6mm}
    \captionof{figure}{
    \textbf{Visualization of the attention score.} It shows that MoCha can automatically trace the character with the mask of only one frame.}
    \label{fig:attention}
    \vspace{-1.5em}
\end{figure}
\textbf{Tracking ability of video diffusion model.}~
Our method, MoCha, leverages the inherent tracking capabilities of video diffusion models, requiring only a single-frame mask to identify the target character, rather than a mask sequence for the entire video. 
To validate this tracking ability, we visualize the attention score map between the mask latent and the source video latent. 
As shown in Fig.~\ref{fig:attention}, the regions corresponding to the selected character consistently maintain high attention scores across different frames.
This observation confirms that the model can effectively track the specified subject throughout the video sequence, obviating the need for sequential masks.

\begin{figure}
    \centering
    \includegraphics[width=\linewidth]{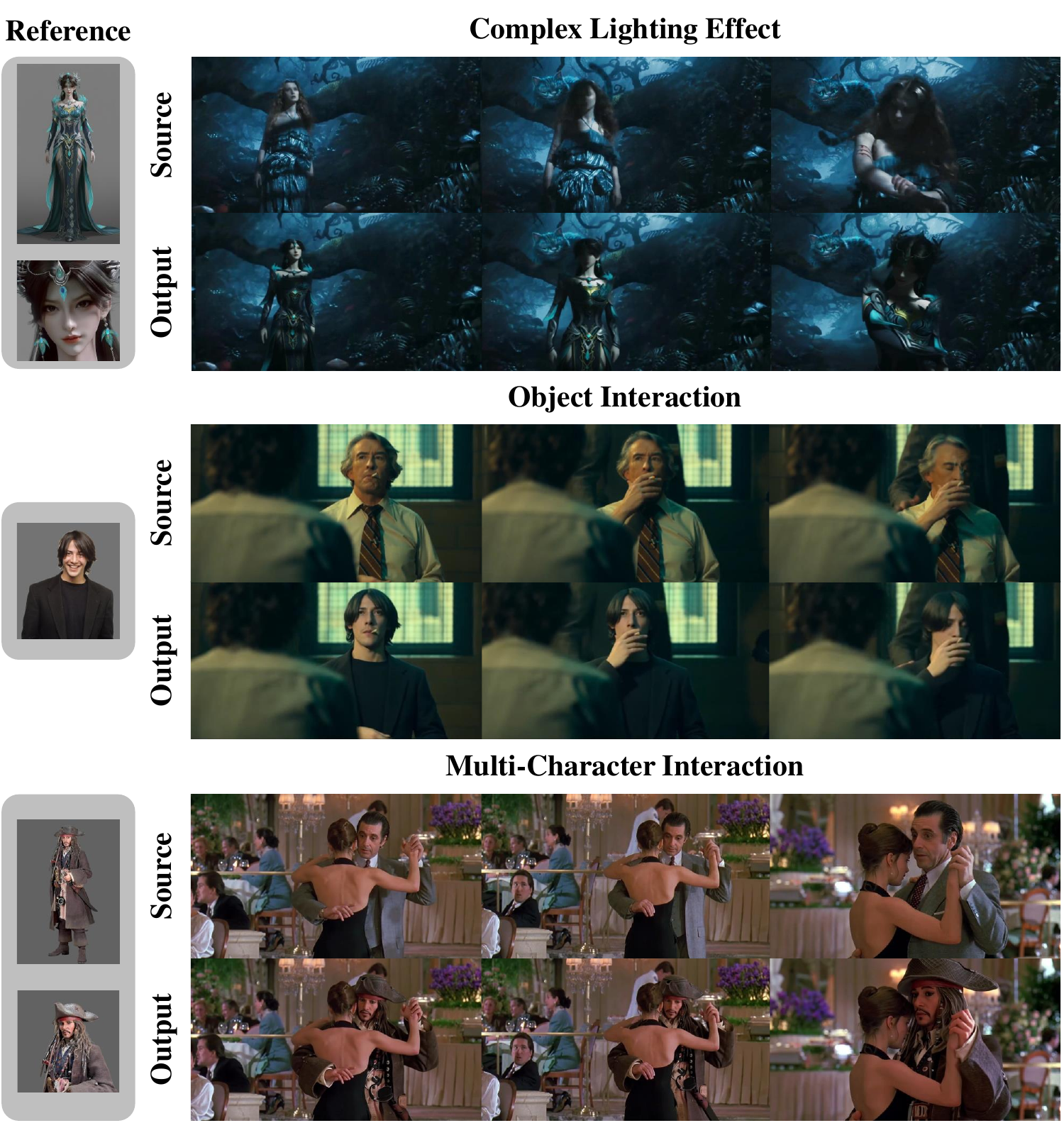}
    \captionof{figure}{
    \textbf{MoCha's performance on complex scenarios.} MoCha demonstrates its robustness and practical utility on complex cases that are commonly encountered in real-world film and video production.}
    \label{fig:complex_case}
    \vspace{-1.5em}
\end{figure}
\medskip
\noindent\textbf{Performance on Complex Cases.} 
To demonstrate the wide applicability and practical utility of MoCha, we evaluate its performance using a collection of more complex test cases. We focus on challenging examples by seeking difficult scenarios across three distinct dimensions: 1) Complex lighting effect. 2) Object interaction. 3) Multiple-character interaction. Fig.~\ref{fig:complex_case} illustrates MoCha's performance in these scenarios. The results confirm that our model consistently generates highly realistic and high-quality video even under these conditions.

\medskip
\noindent\textbf{Application beyond Character Replacement.}
Although our method is primarily developed for the character replacement task, we observe that MoCha demonstrates the capability to handle other types of subject replacement and video editing. Specifically, our model can seamlessly replace non-human subjects from the source video. Furthermore, by editing the reference character's face and clothing using image-editing models, MoCha can be readily applied to tasks such as face swapping and virtual try-on. Some novel and illustrative results showing this expanded utility are presented in the supplementary material. In summary, these findings highlight MoCha's strong generalization ability and broad practical utility.

\section{Conclusion}
In this paper, we introduce MoCha, the first end-to-end video character replacement framework. 
Distinct from current reconstruction-based methods that rely on dense guidance, MoCha efficiently fulfills the replacement task, requiring only a single arbitrary frame mask by harnessing the tracking ability of the video diffusion model.
MoCha faithfully transfers original character motion and facial expression through efficient in-context learning. 
To effectively adapt the multi-modal input condition and enhance facial identity, we introduce a condition-aware RoPE and an RL-based post-training stage guided by a differentiable facial function.
Extensive experimental results demonstrate that MoCha significantly outperforms existing methods.
Finally, we show that MoCha exhibits wide applicability beyond video character replacement, extending to applications such as face swapping and virtual try-on.

{
    \small
    \bibliographystyle{ieeenat_fullname}
    \bibliography{main}
}


\end{document}